# Jabalín: a Comprehensive Computational Model of Modern Standard Arabic Verbal Morphology Based on Traditional Arabic Prosody

Alicia González Martínez[1], Susana López Hervás[1], Doaa Samy[2],
Carlos G. Arques[3]†*, Antonio Moreno Sandoval[1]†*
[1]LLI-UAM, Universidad Autónoma de Madrid
{a.gonzalez,antonio.msandoval}@uam.es, hervas.susana@gmail.com
[2]Spanish language Department, Cairo University
doaasamy@cu.edu.eg
[3] Dept. of Development & differentiation,
Spanish National Center for Molecular Biology (CBM-SO)
cgarcia@cbm.uam.es

† These authors contributed equally to this work.
* Corresponding authors

**Abstract.** The computational handling of Modern Standard Arabic is a challenge in the field of natural language processing due to its highly rich morphology. However, several authors have pointed out that the Arabic morphological system is in fact extremely regular. The existing Arabic morphological analyzers have exploited this regularity to variable extent, yet we believe there is still some scope for improvement. Taking inspiration in traditional Arabic prosody, we have designed and implemented a compact and simple morphological system which in our opinion takes further advantage of the regularities encountered in the Arabic morphological system. The output of the system is a large-scale lexicon of inflected forms that has subsequently been used to create an Online Interface for a morphological analyzer of Arabic verbs. The Jabalín Online Interface is available at http://elvira.lllf.uam.es/jabalin/, hosted at the LLI-UAM lab. The generation system is also available under a GNU GPL 3 license.

## 1 Introduction

Morphological resources are essential components of more complicated systems used in domains such as artificial intelligence, automatic translation or speech recognition systems. Thus, the quality of the resource will strongly affect the whole system. This makes it crucial to develop robust and comprehensive morphological applications.

In the field of Arabic language processing, the existing models have exploited the regularities encountered in the language to a variable extent, yet we believe there is still some scope for improvement. We intend to fill this gap developing a robust and compact system that covers all Arabic verbal morphology by means of simple and general procedures.

Modern Standard Arabic (MSA) is the formal language most widely used nowadays in the whole Arab world. It is spoken across more than 20 countries by over 300 million speakers [1]. MSA stands out for being the language of the media, and in general it is used in all formal situations within society. It is also the language of higher education, and it is used in most written texts. MSA is not a natural language, since it does not have real native speakers [2, 3, 4]. The native languages of Arab people are what we call the Arabic spoken varieties—they learn MSA through the educational system, thus in a non-natural way.

As it is not a natural language, MSA morphology has been described as an extremely regular system [5], susceptible of being represented by means of precise formal devices. As Kaye describes it, MSA presents an "almost (too perfect) algebraic-looking grammar" [2:665]. Broadly speaking, stems—word-forms without the affixal material [6]—are generally built by the organized combination of two types of morphemes—what we call the *root* and the *pattern*. The MSA lexicon contains between 4000 to 5000 different roots [7,8], and verbal morphology exhibits 24 different patterns, of which 16 are really common. Semantically related words tend to share the same root morpheme. Thus, the root turned out to be the basic component of Arabic lexicography, to the extent that dictionaries are organized by roots [9]. At a more superficial level, the inflectional system applies several operations to turn stems into specific verbal wordforms. This stage is considerably complicated by the interaction of phonological and orthographic alterations. All these phenomena hinder the process of formalizing the sytem, thus making it an extremely interesting and challenging task.

## 1.1 MSA Morphotactics

MSA presents two morphological strategies: *concatenative* and *non-concatenative*—also known as *templatic morphology*. Concatenative morphemes are discrete segments which are simply attached to the stem regardless of the position, i.e., they have the form of an uninterrupted string. Non-concatenative morphemes are interleaved elements inserted within a word—they do not form a compact unit, but a discontinuous string whose 'internal blanks' are filled out with other morphemes. In MSA, derivational morphology is mainly marked by non-concatenative schemes, whereas inflectional morphology tends to be concatenative.

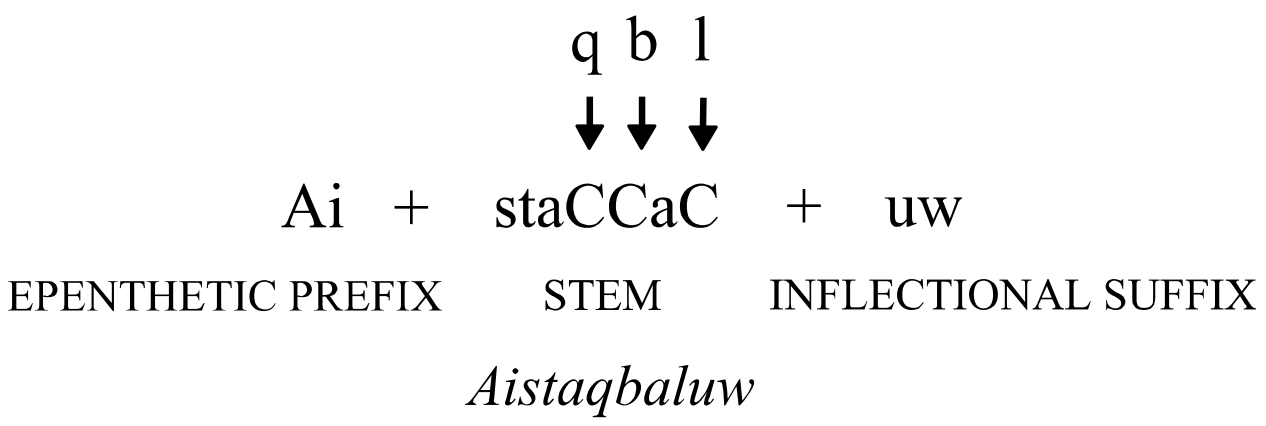

**Fig.** 1. Example of concatenative and non-concatenative processes in the formation of the verbal wordform إستقبلوا *Aistaqbaluw* 'they received'.

Templatic morphology is known in the field of Arabic linguistics as *root-and-pattern morphology*. It takes its name from the Arabic morphemes which have a non-concatenative shape: the root and the pattern. This theoretical description attemps to describe how Arabic stems are built—root-and-pattern morphology states that stems are composed by these two elements. A *root* is a decomposable morpheme that provides the basic meaning of a word, and generally consists of 3 or 4 ordered consonants in non-linear position within the word [10,11,12,13,1]. The *pattern* is a syllabic structure which contains vowels, and sometimes consonants, in which the consonants of the root are inserted and occupy specified places [14,15]. Thus, by the interdigitation of a root and a pattern stems are created [16,17,18,10,15]. Some authors have proposed to separate the vowels from the template and to consider it a separate morpheme. This morpheme is commonly known as *vocalism* [19,20,21,22].

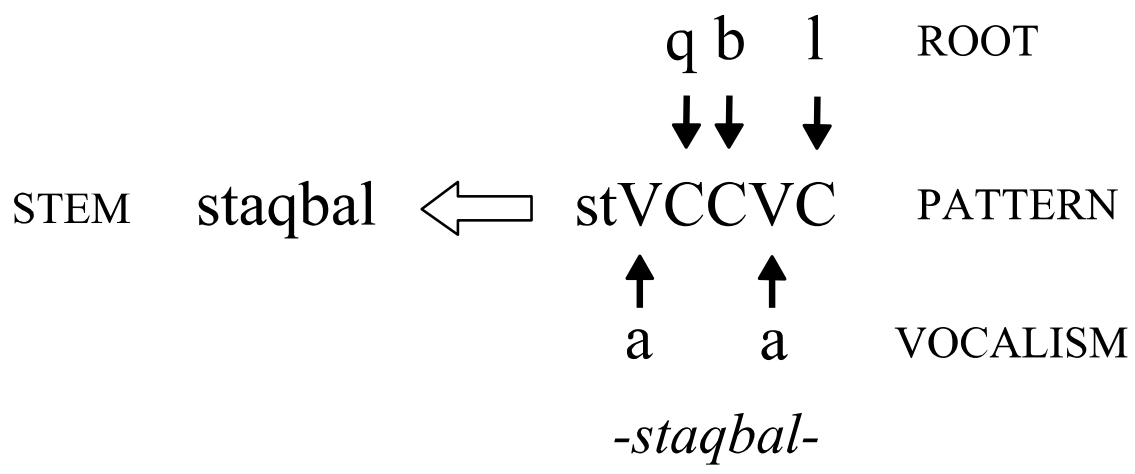

**Fig.** 1. Decomposition of the stem -*staqbal*- from the verbal wordform إستقبلوا *Aistaqbaluw* 'they received' into root, vocalism and pattern.

## 1.2 MSA verbal system

MSA exhibits 24 different verbal patterns. Some of them belong in fact to Classical Arabic and are rarely used. Traditionally they are classified in patterns from 3-consonant roots and patterns from 4-consonant roots. The different patterns add extensions to the basic meaning expressed by the root, i.e., they are of derivational nature. Below, we include the list of patterns using the root فعل *fçl* 'doing'. This root is traditionally used in Arabic to refer to grammatical forms. Patterns are shown using the lemma of the verb, which corresponds to the third person masculine singular of the perfective active inflection [4,10,18,23,24].

Following the Arabic western linguistic tradition, we use Roman numerals to refer to the different patterns. Patterns I include two vowels in their specification: one corresponds to the thematic vowel of the perfective and the other one to the thematic vowel of the imperfective—both correspond to the second vowel position of the stem. Some verbs share the same lemma form, but they are considered different since they present different forms in their conjugation. 4-consonant roots are distinguished from 3-consonant roots by the addition of a ‘Q’ to the Roman numeral.

**Table 1.** List of all verbal patterns in Arabic. Information on the transliteration system used throughout the whole paper can be found at http://elvira.lllf.uam.es/ING/transJab.html.

| **Pattern** | **lemma from root فعل *fçl*** | **example** |
|---|---|---|
| Iau | فَعَلَ façala | كَتَبَ *kataba* ‘to write’ |
| Iai | فَعَلَ façala | رَمَى *ramaY* ‘to throw’ |
| Iaa | فَعَلَ façala | ضَرَبَ *Daraba* ‘to hit’ |
| Iuu | فَعُلَ façula | كَبُرَ *kabura* ‘to grow’ |
| Iia | فَعِلَ façila | رَضِيَ *raDiya* ‘to agree’ |
| Iii | فَعِلَ façila | دَرِثَ *wariþa* ‘inherit’ |
| II | فَعَّلَ faç~ala | عَلَّمَ *çal~ama* ‘to teach’ |
| III | فَاعَلَ faAçala | شَاهَدَ *XaAhada* ‘to watch’ |
| IV | أَفْعَلَ Áafçala | أَحَبَّ *ÁHab~a* ‘to love’ |
| V | تَفَعَّلَ tafaç~ala | تَعَلَّمَ *taçal~ama* ‘to learn’ |
| VI | تَفَاعَلَ tafaAçala | تَآمَرَ *taÃmara* ‘to plot’ |
| VII | اِنْفَعَلَ Ain·façala | اِنْقَضَى *Ain·qaDaY* ‘to pass’ |
| VIII | اِفْتَعَلَ Aif·taçala | اِتَّفَقَ *Ait~afaqa* ‘to agree’ |
| IX | اِفْعَلَّ Aif·çalla | اِحْمَرَّ *AiH·mar~a* ‘to turn red’ |
| X | اِسْتَفعَلَ Aistaf·çala | اِسْتَمَرَّ *Ais·tamar~a* ‘to continue’ |
| XI | اِفْعَالَّ Aif·çaAlla | اِحْمَارّ *AiH·maAr~a* ‘to tuen red’ |
| XII | اِفْعَوْعَلَAif·çaw·çala | اِحْضَوْضَرَ *AiHDawDara* ‘to become green’ |
| XIII | اِفْعَوَّلَ Aif·çaw~ala | اِجْلَوَّذَ *Aijlaw~aða* ‘to last long’ |
| XIV | اِفْعَنْلَلَ Aif·çan·lala | اِسْحَنْكَكَ *Ais·Han·kaka* ‘to be dark’ |
| XV | اِفْعَنْلَى Aif·çan·laA | اِعْلَنْدَى *Aiç·lan·daY* ‘to be stout’ |
| QI | فَعْلَلَ faç·lala | تَرْجَمَ *tar·jama* ‘to translate’ |
| QII | تَفَعْلَلَ tafaç·lala | تَدَحْرَجَ *tadaH·raja* ‘to roll’ |
| QIII | اِفْعَنْلَلَ Aif·çanlala | اِسْلَنْطَحَ *Ai·slan·TaHa* ‘to lie on one’s face’ |
| QIV | اِفْعَلَلَّ Aif·çalal~a | اِقْشَعَرَّ *Aiq·Xaçar~a* ‘to shudder with horror’ |

Regardless the pattern, each verb may present a full conjugational paradigm. The paradigm exhibits a tense/aspect marking, opposing perfective and imperfective. Imperfective, in turn, includes three possible moods: indicative, subjunctive and jussive. There is an imperative conjugation, derived from the imperfective form. At the same time verbs exhibit voice opposition in active and passive, which consists only in a different vocalization. Each conjugational paradigm shows the features of person, number and gender [4,10,18,23,24]. Obviously, verbs do not cover all the possible inflectional alternatives. In the following table, we can see the full conjugational paradigm of the active verb فَعَلَ *façala* ‘to do’.

**Table 1**. Complete conjugational paradigm of the Arabic active verb فَعَلَ *façala* 'to do'. The information of the inflectional tag is as follows. First position: 1=first person; 2=second person; 3=third person. Second position: S=singular; D=dual; P=plural. Third position: M=masculine; F=feminine; N=non-marked for gender.

| Inflec Tag | Perfective | Imperfective | | | Imperative |
|---|---|---|---|---|---|
| | | Nominative | Subjunctive | Jussive | |
| 3SM | فَعَلَ façala | يَفْعُلُ yaf·çulu | يَفْعُلَ yaf·çula | يَفْعُلْ yaf·çul· | |
| 3SF | فَعَلَتْ façalat· | تَفْعُلُ taf·çulu | تَفْعُلَ taf·çula | تَفْعُلْ taf·çul· | |
| 3DM | فَعَلَا façalaA | يَفْعُلَانِ yaf·çulaAni | يَفْعُلَا yaf·çulaA | يَفْعُلَا yaf·çulaA | |
| 3DF | فَعَلَتَا façalataA | تَفْعُلَانِ taf·çulaAni | تَفْعُلَا taf·çulaA | تَفْعُلَا taf·çulaA | |
| 3PM | فَعَلُوا façaluwA | يَفْعُلُون! yaf·çuluwna | يَفْعُلُوا yaf·çuluwA | يَفْعُلُوا yaf·çuluwA | |
| 3PF | فَعَلْنَ façal·na | يَفْعُلْنَ yaf·çul·na | يَفْعُلْنَ yaf·çul·na | يَفْعُلْنَ yaf·çul·na | |
| 2SM | فَعَلْتَ façal·ta | تَفْعُلُ taf·çulu | تَفْعُلَ taf·çula | تَفْعُلْ taf·çul· | أُفْعُلْ Auf·çul· |
| 2SF | فَعَلْتِ façal·ti | تَفْعُلِينَ taf·çuliyna | تَفْعُلِي taf·çuliy | تَفْعُلِي taf·çuliy | أُفْعُلِي Auf·çuliy |
| 2DN | فَعَلْتُمَا façal·tumaA | تَفْعُلَانِ taf·çulaAni | تَفْعُلَا taf·çulaA | تَفْعُلَا taf·çulaA | أُفْعُلَا Auf·çulaA |
| 2PM | فَعَلْتُمْ façal·tum | تَفْعُلُونَ taf·çuluwna | تَفْعُلُوا taf·çuluwA | تَفْعُلُوا taf·çuluwA | أُفْعُلُوا Auf·çuluwA |
| 2PF | فَعَلْتُنَّ façaltun~a | تَفْعُلْنَ tafçul˙na | تَفْعُلْنَ tafçulna | تَفْعُلْنَ tafçulna | أُفْعُلْنَ fçul·na |
| 1SN | فَعَلْتُ façal·tu | أَفْعُلُ Áaf·çulu | أَفْعُلَ Áaf·çula | أَفْعُلْ Áaf·çul· | |
| 1PN | فَعَلْنَا façal·naA | نَفْعُلُ naf·çulu | نَفْعُلَ naf·çula | نَفْعُلْ naf·çul· | |

At a superficial level, the whole verbal system is highly affected by allomorphism. Allomorphism is the situation in which the same morpheme exhibits different phonological shapes depending on the context [6]. This determines that a set of surface representations can be related to a single underlying representation [6]. Allomorphism is one of the most complicated aspects of Arabic morphological analysis.

The main causes of allomorphism in MSA are phonological constraints on the semiconsonants *w* and *y*. Verbs with roots containing at least one semiconsonant phoneme typically present phonological alterations. Another cause of allomorphism is the presence of two identical consonants in the second and third place of the root, which is known as *geminated or doubled roots* [4,10,23]. In spite of the uniform nature of these phonological alterations, which are susceptible to systematization, verbs suffering these phenomena are considered irregular in traditional Arabic grammar.

Orthographic idiosyncracies are closely related with these phonological alterations. Thus, we can refer to them as *orthographic allomorphism*. Alhough not having linguistic nature, they are relevant computationally.

## 1.3 Traditional Arabic Prosody

Medieval Arab scholars developed an interesting analysis of Arabic morphological structure. With the development of Arabic poetry, scholars noticed that Arabic prosodic units were subjected to a marked rhythmic uniformity. This may be partially due to the fact that Arabic phonotactics restricts many types of syllables. Essentially, MSA accepts three types: CV, CVC and CVV. Exceptionally CVVC and CVCC are permitted [24].

The most important contribution in this field was made by Al Khalil, an acclaimed Arab scholar considered the father of Arabic lexicography. He described and systematized the metrical system of Arabic poetry, based directly on orthography. One of the interesting things of the Arabic writing system is that only consonants are considered letters. Vowels are diacritic symbols written on or below the consonant they accompany. In order to define the different metrical patterns, Al Khalil classified letters in two types [25]:

*1. sakin* letter حرف ساكن 'static letter', i.e. an unvocalized letter. A consonant without a vowel, or a semiconsonant. It is important to note that semiconsonants are used to represent long vowels when preceded by a short vowel.

*2. mutaharrik* letter حرف متحرك 'moving letter', i.e. a vocalized letter. A *mutaharrik* letter is a consonant followed by a diacritic vowel.

The fundamental principle of the analysis of al-Khalil is that a *mutaharrik* letter is heavier than a *sakin*. To represent this, they are marked with different weight symbols. A mutaharrik letter is going to be assigned the value 1, and a sakin letter the value of 0[1]. Thus, an orthographic word can be represented as 1-0 combinations. These combinations are subsequently classified in wider groups of weight, which in fact unravel the different syllabic structures. First, 1-0 combinations compute the value 2; then 1-2 combinations compute the value 3; at last, 2-2 combinations compute the value 4. Finally, we can sum the resulting numbers and get the total weight for a word. Below, this computation of lexical weight is shown.

| | | | | | |
|---|---|---|---|---|---|
| Word | يُعَلِّمُ *yuçal~imu* 'he teaches' | | | | |
| Letters segmentation | yu | ça | l | li | mu |
| Weights | 1 | 1 | 0 | 1 | 1 |
| Accumulative weights | 1 | 2 | | 1 | 1 |
| | | 3 | | 1 | 1 |
| Total weight of lexical item | | | 5 | | |

The fact that a small number of syllabic structures is allowed by Arabic phonotactics has interesting implications: as the formation of words belonging to the same morphological class is the product of a *quasi* mathematical combination of similar morphemic material, the resulting syllabic structure will tend to follow the same patterns. Thus, it seems possible to propose a precise formalism which predicts the syllabic structures for Arabic lexical items.

## 1.4 Current Computational Systems of MSA Morphology

The aim of Natutal Language Processing (NLP) is to find the most efficient way to describe formally a language for a specific application. The core task in this field is to build a morphological analyzer and generator. Morphological analyzers are composed of two basic parts [21]:

1. Lexical units , i.e, a lexicon responsible for the coverage of the system. Ideally, the lexicon should include all the morphemes of a given language.

2. Morphosyntactic knowledge, i.e, a set of linguistic rules responsible for the robustness of the system. There are mainly two types of rules:

   (a) rewrite rules, which handle the phonological and orthographic variations of the lexical items;

   (b) morphotactic rules, which determine how morphemes are combined.

In fact both the lexicon and the rule components are closely related: linguistic rules can be codified in the lexicon, and consequently the size of both parts is directly related.

An early implementation of a computational model of Arabic morphology was carried out by Kenneth Beesley [8,15,16]. He created the Xerox Arabic morphological Analyzer, which uses finite-state technology to modelize MSA morphology. Beesley created a separate lexicon for each morpheme type: prefixes, suffixes, roots (4,930 entries) and patterns (about 400 entries). Information on root and pattern combinations are stored in the lexicon of roots, so he included full phonotactic coding in the entries. The system extracts the information stored in the lexicons and compiled it into a finite state transducer (FST). Phonotactics and orthographic variation rules are also compiled into FSTs. The combination of prefixes, stems and suffixes yields over 72 million hypothetical forms—with the disadvantage that it overgenerates. The phonotactic treatment includes 66 finite-state variation rules.

Beesley's system presents a fairly elegant description of MSA morphology. On the negative side, he uses an extensive list of patterns, as it is common in the traditional descriptions of Arabic morphology.

The most famous analyzer for the Arabic language is the Standard Arabic Morphological Analyzer (SAMA), formerly known as Buckwalter Arabic Morphological Analyzer (BAMA)—up to version 3—which was created by Tim Buckwalter in 2002 [1,26]. It has become the standard tool in the field of Arabic NLP [27]. SAMA is strongly lexicon-based—Buckwalter sacrifices the possibility of using a linguistic model in favour of a very practical solution: he codifies all linguistic processes in the lexicon and uses the stem as the basic lexical entry. He then specifies various sets

1 In Arabic, the letter hamza ه is used to represent the *sakin* letter and the numeral ١ for the *mutaharrik* [25].

of rules based on concatenative procedures to establish the permitted combinations of the different lexical items. The lexicon of stems includes 79,318 entries, representing 40,654 lemmas. Stems are turned into underlying forms by the addition of affixes, compiled in a lexicon of prefixes (1,328 entries) and a lexicon of suffixes (945 entries) These lexicons include both affixes and clitics.

This system presents two important disadvantages: first, it has a lot of obsolete words, reducing considerably its efficiency [27, 28]. Attia estimates that about 25% of the lexical items included in SAMA are outdated [27]. Second, it does not follow a linguistic analysis of MSA morphology. The design of morphology implies that phonological, morphological and orthographic alterations are simply codified in the lexicon: the same word may have more than one entry in the lexicon according to the number of lexemes its inflectional set of forms presents.

A very recent analyzer is the AraComLex, a large-scale finite-state morphological analyzer toolkit for MSA developed principally by Mohammed Attia[14,27]. Its lexical database uses a corpus of contemporary MSA to reject outdated words. It also makes use of pre-annotation tools and machine learning techniques, as well as knowledge-based pattern matching, to automatically acquire lexical knowledge. As a result, AraComLex is the only Arabic morphological analyzer which includes strictly contemporary vocabulary and is highly enhanced with additional linguistic features. Attia chooses the lemma as the basic lexical entry. The lexicon of lemmas has 5,925 nominals, 1,529 verbs; the lexicon of patterns 456 nominal patterns and 34 verbal. There are 130 alteration rules to handle all alterations encountered in the lexicon. Attia notes that a stem-based system, like that of the SAMA, is more costly for it has to list all the stem variants of a form, whereas a lexeme-based system simply includes one entry for each lexical form and a set of generalized rules for handling the variations. He also rejects a root-based approach, as it is more complex and tends to cause overgeneration problems.

The AraComLex is possibly the most consistent morphological analyzer for MSA, not only for its acccuracy and efficient implementation, but also for its ease of use—and gladly it is available under a GNU GPL licence. However, it did not intend to present a comprehensive model of Arabic internal morphology.

## 2 Methodology

The computational system has been implemented in Python programming language (version 3.2). In recent years it has come to be one of the best options for developing applications in the field of NLP. Further, version 3 of Python fully supports Unicode, so it can directly handle Arabic script. In relation to orthography, we handle fully diacritized forms. Arabic uses diacritics to disambiguate words [29], and thus we keep them to create a lexicon as unambiguous as possible. The rules of phonotactic and orthotactic nature, which cause a gap between the underlying—regularized—form and the surface form, were formalized using regular expressions.

We have manually created a lexicon of Arabic verb lemmas which consists of 15,453 entries with unambiguous information of each verbal item. The lexicon will be used as an input for the system of verbal generation. It was taken from a list of verbs included in the book *A dictionary of Arabic verb conjugation* by Antoine El-Dahdah [30]. The lexicographical sources used by El-Dahdah to compose his lexicon are widely known classical dictionaries. Thus, the lexicon includes many outdated vocabulary. Although this is a drawback for the developmet of a practical and accurate resource, this is going to allow us to have a complete overview of the MSA verbal system.

## 3 Results

Based on the ideas of Arabic traditional prosody, we have designed and built a computational model that describes the MSA verbal system. The computational model is based on generation. The output of the system is a large-scale lexicon of fully diacritized inflected forms. The lexicon has been subsequently used to develop an online interface of a morphological analyzer for verbs.

### 3.1 The Design of MSA Verbal Morphology

Our first aim was to clearly separate morphological phenomena from phonological and orthographic operations. We noticed that all verbs, regardless their nature, can be generated as regular, and then subjected to the constraints of phonology and orthography. By doing so, we can describe a completely regular morphology, applicable to all verbs. In a superficial level, phonological and orthographic alterations can be applied to these regular forms so they get their real superface form. This allows us to focus on morphological traits independently.

At a deep level, we decompose the stem into four elements: a root, derivational processes—consisting mainly of the insertion of consonantal material—a vocalism and a template.

The root is specific for each verb. As we have already said, it consists of three or four consonants interdigitated thoughout the verbal stem. For instance, the root of the verb أرسل *Ársala* 'to send' is رسل *rsl*. There are cases, however, in which the root is not transparent, as in the verb استجاب *AistajaAba* 'to respond', whose root is جوب *jwb*.

The derivational processes—which correspond to parts of the traditional patterns—tend to add semantic connotations to the basic sense of the verb's root. The processes consist of three types of operations:

1. Insertion of one or more consonants, as the affix 'st' in the verb استجاب *AistajaAba* 'to respond'.

2. Insertion of a vowel lengthening mark, as the element *A* in the verb شاهد *XaAhada* 'to watch' which, is used to lengthen the vowel *a*.

3. Duplication of a consonant, as in the verb علّم *çal~ama* 'to teach', which doubles the *l*. The symbol ~ is used in the transliteration to represent the Arabic character ّ whose function is to double the sound of a consonant.

In MSA there are only three short syllables *a*, *i* and *u*. The vocalism morpheme, which consists of two vowels—a first vowel and a second vowel within the stem—just presents different combinations of vowels in the vocalic slots of the template. For instance, in the inflected form يُرسِلُ *y-ursil-u* 'he sends', the stem shows two vowels, *u* and *i*. In this case, the vowels depend on the form of the stem, i.e., they have a default content, but in other cases they must be marked lexically.

The template is the most interesting element in the formation of the stem, for it has to deal with the combination of all the previous elements. This lead us to the challenging task of devising an algorithm that specifies how the lexical items are combined and merged into a well defined template.

We stated that we believe that the syllabic skeleton of Arabic verbal stems can be formalized in a reduced set of basic structural units. We relied this hypothesis on al-Khalil's works on quantitative prosodic theory, for it computes syllabic shape by means of a systematic and simple mathematical device based on orthography. Al-Khalil's counting procedure hints at the existence of an extremely regular system of syllabic structure in Arabic. The interesting thing here is that verbs belonging to the same morphological class, overwhelmingly show the same weight, regardless being classified as regular or irregular.

Following this idea, we established that templates are formed by two basic units: first, consonants and vowel lengthening elements, and second, vowels themselves.We refer to the former as F, and to the latter as V or W (for first and second vowel respectively). A detailed analysis led us to propose that there are only two types of templates which cover all the traditional verbal patterns in the Arabic system. The basic difference between these two types is the length of the penultimate syllable: on one type this syllable is heavy, and on the other it is light. Hence, we are going to name the first type *H*, for *heavy,* and the second *L*, for *light*. Both types distinguish a perfective stem (*p-stem*), an imperfective stem (*i-stem*), and an imperative stem (*m-stem*), as each verb presents these three stems along its conjugation[2].

**Table** 1. Classification of verbal templates.

| Template type | p-stem | i-stem | m-stem |
|---|---|---|---|
| L | FFVFWF | VFFFWF | FFFWF |
| H | FFVFFWF | VFFFFWF | FFFFWF |

The algorithm for combining the lexical items and the template is quite simple. First, the root and the derivational material are merged to form a string. This string is inserted into the template by a simple procedure. Each character from the root plus derivation string replaces an F of the template, starting from the end. If there are some F slots left after the replacement process, they are removed from the resulting string. Then, the specific vowels of the stem replace the V and W symbols. This straightforward algorithm is shown in figure 3. Strikingly, this algorithm implies that verbs

2 m-stem is actually the same as i-stem but without the first vowel. For the sake of simplicity, we preferred to keep it as an autonomous template.

of 3-consonant and 4-consonant roots are treated the same, so we do not need to have different conjugational categories for them, as is the general custom.

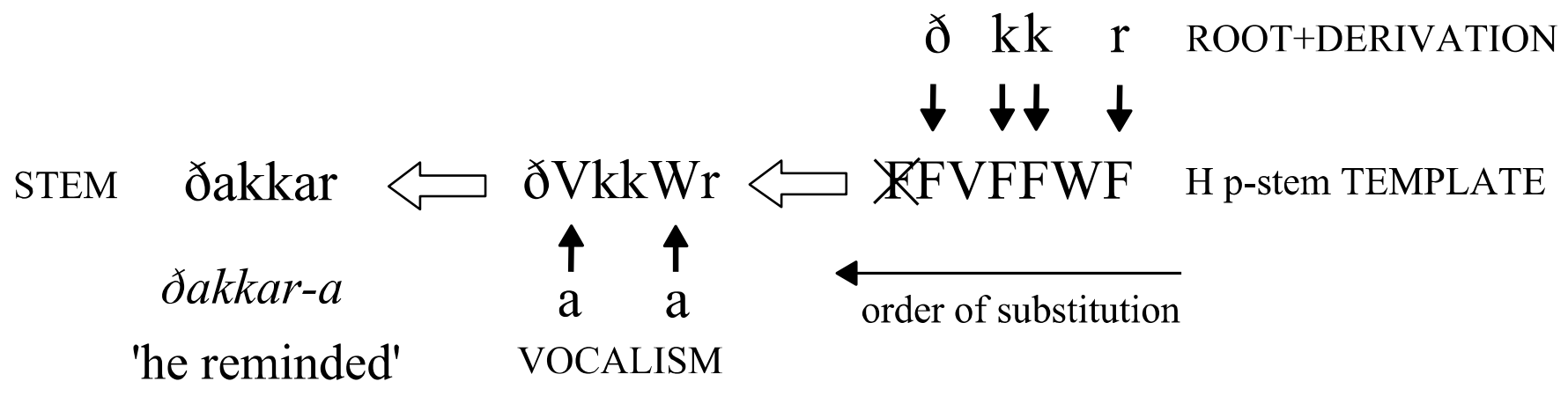

**Fig. 1.** Algorithm for template adjustment. Example of wordform ذَكَّرَ *ðakkara* 'he reminded', root ذكر *ðkr.*

As for the conjugational paradigm, we simply defined the inflectional morphemes that must be added to a base stem so that it turns into an inflected wordform. The whole inflectional paradigm can be seen in Table 4.

**Table 1.** Inflectional Chart. Symbol 'E' represents vowel lengthening.

| TAG | p-stem | i-stem | | | | m-stem |
|---|---|---|---|---|---|---|
| | | all | indicative | subjunctive | jussive | |
| **1SN** | -تُ | أـ | ـُ | ـَ | - | *None* |
| **1PN** | E-نَ | نـ | ـُ | ـَ | - | *None* |
| **2SM** | - تَ | تـ | ـُ | ـَ | - | - |
| **2SF** | -تِ | E تـِ | -نَ | - | - | E-ِ |
| **2DN** | E-تُمَ | E تـَ | -نِ | - | - | E-َ |
| **2PM** | -تُم | E تـُ | -نَ | -ا | -ا | اE-ُ |
| **2PF** | -تُنَّ | تـ-نَ | - | - | - | -نَ |
| **3SM** | -َ | يـ | ـُ | ـَ | - | *None* |
| **3SF** | -تَ | تـ | ـُ | ـَ | - | *None* |
| **3DM** | E-َ | E يـَ | -نِ | - | - | *None* |
| **3DF** | E-َتَ | E تـَ | -نِ | - | - | *None* |
| **3PM** | E-ُ | E يـُ | -نَ | -ا | -ا | *None* |
| **3PF** | -نَ | يـ-نَ | - | - | - | *None* |

In a superficial layer, phonological and orthographic operations modify the underlying form to yield the superficial form of the inflected verb. Even though these phenomena are considered irregular in traditional Arabic grammar, it is essential to note that these alterations are by no means arbitrary, but they entail systematizable subregularities. These operations are formalized as rewrite rules and implemented as regular expressions. The rewrite rules are represented as follows:

> a -> b / _c
>
> *If you find* a *in the word-form, and if* a *is followed by* c*, then change* a *to* b*; where* a *is the pattern we are looking for,* b *is the replacement for the pattern, and* c *is the surrounding context; and the underscore indicates the position of a in relation to c.*

For instance, one of the phonological rules is defined as [uwi -> iy / _Ca]. This rule deals with the sound *wi*, which is a segment discouraged by the Arabic phonological system [31]. Hence, the rule handles the transformation of this sound into a more harmonic sound *iy*. The context specified by the rule indicates that the *pattern* must be followed by a consonant plus a vowel *a*, so that the rule is applied. We can see the behaviour of this rule in the perfective passive formation of the common verb قال *qaAla* 'to say', whose root is *qwl*. By applying this rule, the regularly generated passive **quwila* is substituted for the more melodious sound—and correct—*qiyla.*

In a nutshell, our model is essentially based on the division of stems in a root plus derivational affixation amalgam, a vocalization and a template. These three lexical items are merged by means of a formal device to build verbal stems. The keystone of this procedure are the 2 types of templates and their insertion algorithm, which abstract the syllabic structure of the underlying representation of verbal stems based on predefined basic units.

## 3.2 The Generation Model

The generation system relies on a lexicon of verb lemmas manually compiled for the present project. The sources of this lexicon were described in section 2. Based on our description of verbal morphology, each verb would need two pieces of information: the root, which must be lexically associated to each verb, and a code that codifies the morphemes of the verb stem and its template, i.e., the code shows if the verb presents derivational processes, the vowels it uses for its conjugation and if it adjusts to an L template or to a H template.

The code is formed by six digits and one letter. The latter is placed in position 3 of the code. The meaning of each position is as follows: positions 1, 2 and 4 indicate if the verb has derivational material; position 3 indicates the template the verb follows; and position 5, 6 and 7 indicate the conjugational vowels of the verb. For example, the verb إستمَرَّ *Aistamar~a* 'to continue' has a root مرر and a code 04H0000. The 4 in the code indicates that a prefix 'st' must be adedd to the root, and the H specifies that this verb adjusts to a H template. The zeros indicate that this verb does not have other derivational processes and that the vowels of its conjugation do not have to be lexically marked, i.e., this verbal class has default vowels in its conjugation.

The generation process is as follows. The system generates the conjugation of a verb starting from the verb's root. The code associated to that root is used to keep track of the generation path the verb must follow through the formation of the stem. The system is divided in 7 modules, which follow a hierarchical structure.

Module 1: Root and derivational material merging: in the first stage, the derivational processes are applied to the root. There are 7 processes of consonant insertion, 3 processes of vowel lengthening insertion and 2 process of duplication of a consonant.

Module 2: Insertion into template: the root and derivation amalgam is inserted into the template following the algorithm described in the previous section.

Module 3: Intertion of derivational affix ta- (patterns II and V). We left this single derivational affix to be inserted after the template adjustment for it has a completely different nature, compared to the others. This affix is the only affix constituted by a syllable, contrary to the other affixes, which are single consonants.

Module 4: Insertion of vocalization: vowels are inserted into the template.

Module 5: Phonotactic preprocessing: prohibited syllables are resyllabized and, at this point, deep phonological alterations are carried out—which consist of assimilation processes suffered by forms belonging to the VIII pattern. At the end of this stage the stem if completely formed.

Module 6: Generation of inflectional paradigm: the created stem is passed through the inflectional chart to yield all conjugated forms.

Module 7: Phonotactic constraints and orthographic normalization: all inflected forms are passed through a series of rewrite rules in the form of regular expressions. The rules are hierarchically organized, so if the same form suffers various phonological processes, all are applied in a cascade process. The system has 30 orthographic rules and 33 phonological, making a total of 63 rules to handle verbal allomorphism.

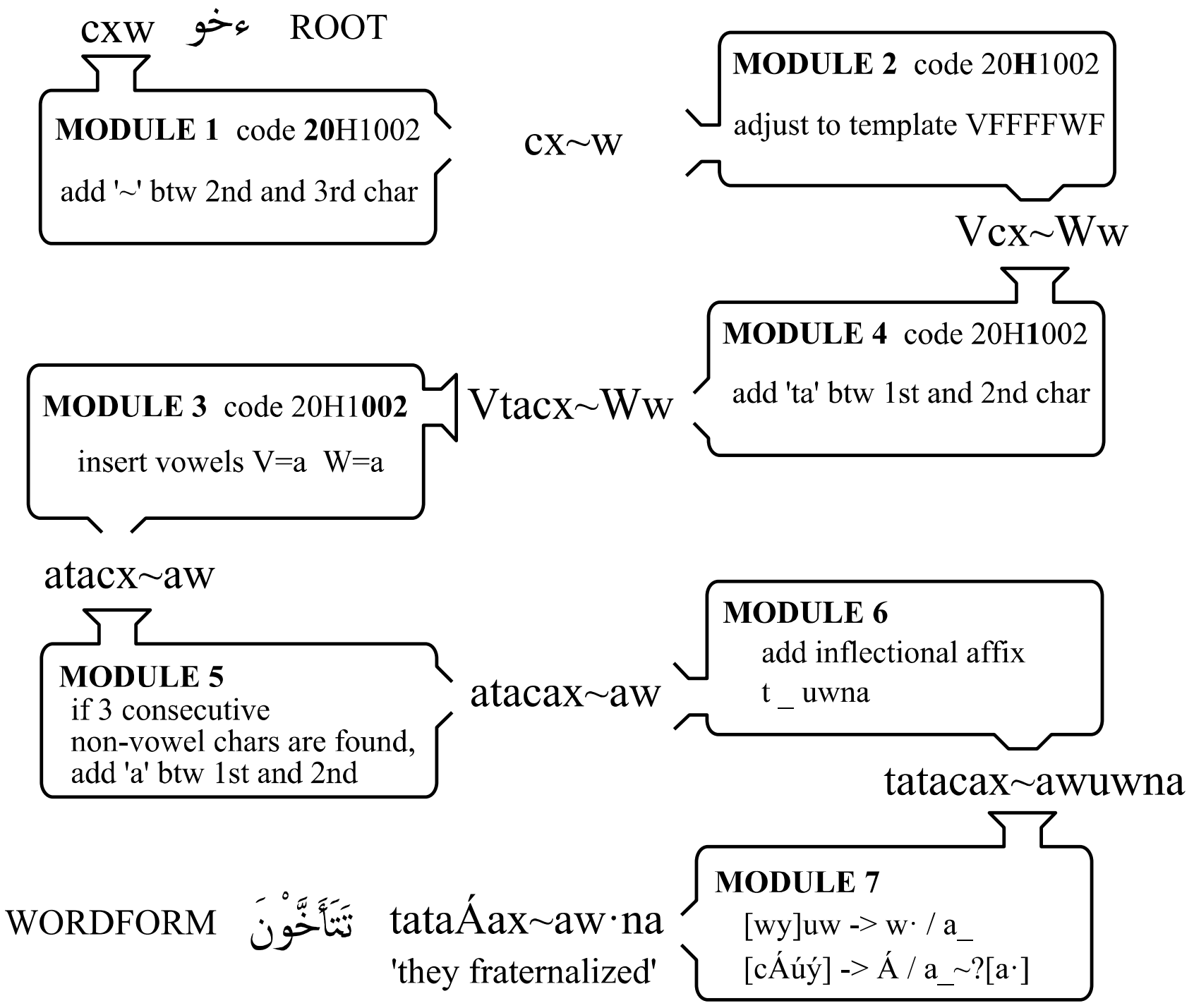

**Fig. 1.** Example of generation of the wordworm تَتَآخَّوْنَ tataÁx~aw·na 'they fraternalized'.

## 3.3 Evaluation of the Model

To evaluate the accuracy of the morphological model, we needed to compare the lexicon generated by our system with a reference lexicon.We carried out the evaluation against the list of inflected verbs extracted from the morphological analyzer ElixirFM [32]. We assumed that the lexicon extracted from the ElixirFM software is a validated database of Arabic conjugation, so we consider it our gold standard. We based this assumption on the fact that ElixirFm is an improvement on the BAMA analyzer, which has reportedly achieved 99.25% precision [33]. Starting from this assumption we normalized the ElixirFm lexicon, so that it shares a common format with our lexicon. In the table below we find the data of both lexicons.

**Table 1.** Data on number of lemmas and forms in ElixirFM and Jabalín

| | No. Lemmas | No. Forms | Forms per lemma |
|---|---|---|---|
| **ElixirFM** | 9009 | 1,752,848 | 192 |
| **Jabalín** | 15,452 | 1,684,268 | 109 |
| **Common** | 6878 | 749,702 (44%) | 109 |

The ElixirFM tagset is redundant, thus the higher number of forms per lemma. Another peculiarity of the ElixirFm tagset is that there may be more than one form corresponding to the same tag. This explains that the total number of forms does not equal the number of lemmas plus the number of forms per lemma.

There are 2,131 lemmas only present in ElixirFM and 8,581 only present in Jabalín. This means that we have a low recall rate with respect to the ElixirFM database. Even though both gaps may seem substantial, we believe that it is an inherent problem of working with Classical Arabic lexicon and, ultimately, both ElixirFM and Jabalín include a high percetage of obsolete lexical entries. There a total of 749,702 common forms. From these, 651 forms were not evaluable because some discrepancies were found in grammar books. This means that the total number of evaluated forms was reduced to 749,051, which represents 44% of our lexicon.

For the evaluation task, we compared the reference lexicon with our generated lexicon and searched for each of our verbal entries in the reference lexicon, obtaining a number of successes and failures. From the evaluable forms, we achieved a precision of 99.52% of correct analyses. We believe that this high accuracy validates our model.

**Table 1.** Results from the evaluation.

| | No. forms | % from total | % from eval |
|---|---|---|---|
| **Correct** | 745,436 | 44,26% | 99.52% |
| **Incorrect** | 3,615 | 0.21% | 0.48% |
| **No data** | 935,217 | 55.53% | - |
| **Total** | 1,684,268 | - | - |

### 3.4 The Jabalín Online Interface

The Jabalín Online Interface is a web application for analyzing and generating Arabic verbs. It uses the lexicon of inflected verbs provided by the generation system described in the previous section. The online interface is hosted at the LLI-UAM laboratory web page, under the address http://elvira.lllf.uam.es/jabalin/.

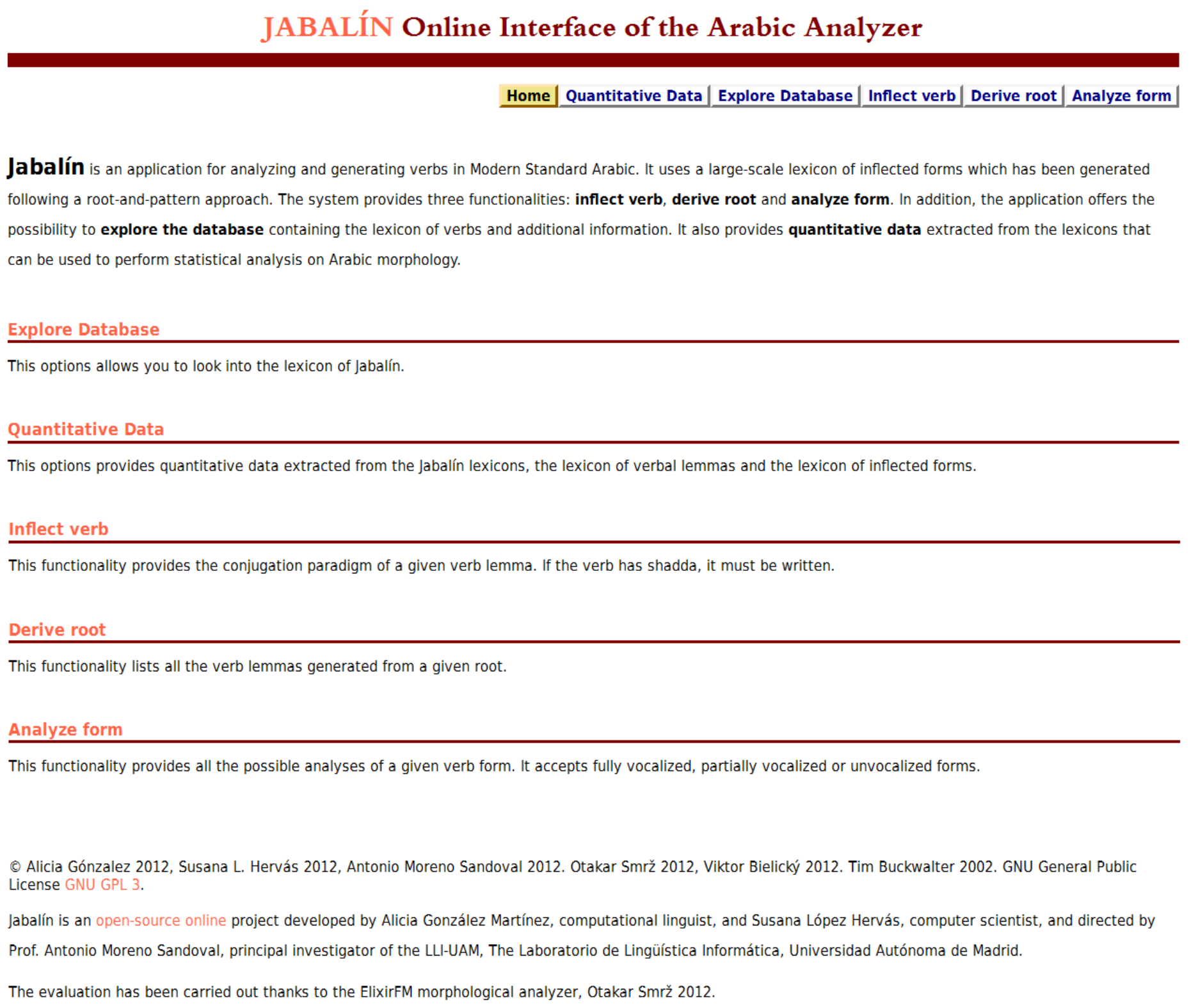

The interface provides five functionalities: explore database, quantitative data, inflect verb, derive root and analyze form. *Explore database* allows one to look into the lexicon of Jabalín. It includes information about all the inflected forms from the lexicon and indicates if the form has been evaluated. *Quantitative data* shows various types of frequency data extracted from both the lexicon of lemmas and the lexicon of inflected forms. *Inflect verb* provides the conjugation paradigm of a given verb lemma, including the root and the pattern of the verb. *Derive root* lists all the verb lemmas generated from a given root and its corresponding patterns. *Analyze form* provides all the possible analyses of a given verbal form. It accepts fully vocalized, partially vocalized or unvocalized forms.

## 4 Conclusions and Future Work

Our model intends to present a compact and efficient implementation of MSA verbal morphology. Our design of morphology is based on a linguistically motivated analysis which takes full advantage of the inner regularities encountered in Arabic morphology.

As a first goal, our descriptive model aims to clearly separate morphological, phonological and orthographic phenomena, avoiding to treat different types of linguistic layers by means of the same operations. One of the keystones of the model is that we present a robust and simple algorithm for dealing with the non-concatenative nature of Arabic morphology, which gave us strikingly good results. As a consequence, we achieved to reduce the traditional classification of Arabic patterns from 24 classes to only 2 conjugational classes. Another remarkable conclusion drawn by the model is that there is no need to morphologically distinguish between verbs of 3-consonant and 4-consonant roots.

We created a total of 63 rules to handle both phonological and orthographic alterations. As a way of testing the robustness of the model, we automatically evaluated 44% of the output lexicon of the system against a gold standard. The results achieved by the evaluation show 99.52% correct forms.

Perhaps, the most remarkable conclusion we take from the template categorization and the ordering algorithm is that Arabic syllabic structure is overwhelmingly regular. The highly restrictive phonotactic system of Arabic makes the syllabic structure of stems predictable. In a nutshell, we have demonstrated that it is possible to develop a precise formalism which predicts the syllabic structures for Arabic lexical items.

As for future works, we strongly believe that in the long run a morphological system based on a precise description of the Arabic morphological system would benefit from high efficiency and better adaptability to numerous applications. Therefore, our forthcoming endeavours will be focused on extending the proposed model to nominal morphology, so that we can develop a complete system to handle Arabic morphology. The nominal system has the disadvantage of being more complex than verbal morphology, yet we believe that the basic principles of our analysis would be maintained in a nominal model.

Furthermore, the efficiency obtained from this system strongly suggests that this descriptional model must have linguistic implications, so one of our most interesting future endeavours is to place this descriptional framework inside current linguistic theory.

**Acknowledgements.** The present work benefited form the input of Jiři Hana and Theophile Ambadiang, who provided valuable comments to the research summarised here. We would like to thank Otakar Smrž too, for his invaluable help regarding the ElixirFM analyzer. This research was supported and funded by an FPU-UAM scholarship from the Universidad Autónoma de Madrid and by a grant from the Spanish Government (R&D National Plan Program TIN2010-20644-C02-03).